\documentclass[runningheads]{llncs}
\usepackage{graphicx}
\usepackage{comment}
\usepackage{amsmath,amssymb} % define this before the line numbering.
\usepackage{color}

\usepackage[utf8]{inputenc} % allow utf-8 input
\usepackage[T1]{fontenc}    % use 8-bit T1 fonts
\usepackage{tabularx}
\usepackage{color}
\usepackage{dsfont}
\usepackage{pifont}% http://ctan.org/pkg/pifont
\usepackage{enumitem}
\usepackage{bm}
\usepackage{cite}
\usepackage{subcaption} % captionsetup
\usepackage{array,multirow} % % multirow
\usepackage{booktabs} % cmidrule
\usepackage{adjustbox}
\usepackage[normalem]{ulem}
\usepackage[para]{footmisc}
\usepackage{pdfpages}

\newcommand{\cmark}{\ding{51}}%
\newcommand{\xmark}{\ding{55}}%
\newcolumntype{L}[1]{>{\raggedright\let\newline\\\arraybackslash\hspace{0pt}}m{#1}}
\newcolumntype{R}[1]{>{\raggedleft\let\newline\\\arraybackslash\hspace{0pt}}m{#1}}
\newcolumntype{C}[1]{>{\centering\let\newline\\\arraybackslash\hspace{0pt}}m{#1}}

\newcommand{\ie}{\textit{i}.\textit{e}.}
\newcommand{\eg}{\textit{e}.\textit{g}.}

\begin{document}
% \linenumbers
\pagestyle{headings}
\mainmatter
\def\ECCVSubNumber{4644}  % Insert your submission number here

\title{Learning with Privileged Information for Efficient Image Super-Resolution} % Replace with your title

\begin{comment}
\titlerunning{ECCV-20 submission ID \ECCVSubNumber} 
\authorrunning{ECCV-20 submission ID \ECCVSubNumber} 
\author{Anonymous ECCV submission}
\institute{Paper ID \ECCVSubNumber}
\end{comment}
%******************

%\begin{comment}
\titlerunning{Learning with PI for Efficient Image SR}
\author{Wonkyung Lee\thanks{equal contribution} \and
Junghyup Lee\inst{*} \and
Dohyung Kim\inst{*} \and
Bumsub Ham\thanks{corresponding author \email{(bumsub.ham@yonsei.ac.kr)}}}
\authorrunning{W. Lee et al.}	
\institute{Yonsei University}
%\end{comment}
%******************
\maketitle

\begin{abstract}
  Convolutional neural networks~(CNNs) have allowed remarkable advances in single image super-resolution~(SISR) over the last decade. Most SR methods based on CNNs have focused on achieving performance gains in terms of quality metrics, such as PSNR and SSIM, over classical approaches. They typically require a large amount of memory and computational units. FSRCNN, consisting of few numbers of convolutional layers, has shown promising results, while using an extremely small number of network parameters. We introduce in this paper a novel distillation framework, consisting of teacher and student networks, that allows to boost the performance of FSRCNN drastically. To this end, we propose to use ground-truth high-resolution~(HR) images as privileged information. The encoder in the teacher learns the degradation process, subsampling of HR images, using an imitation loss. The student and the decoder in the teacher, having the same network architecture as FSRCNN, try to reconstruct HR images. Intermediate features in the decoder, affordable for the student to learn, are transferred to the student through feature distillation. Experimental results on standard benchmarks demonstrate the effectiveness and the generalization ability of our framework, which significantly boosts the performance of FSRCNN as well as other SR methods. Our code and model are available online: \url{https://cvlab.yonsei.ac.kr/projects/PISR}.

\end{abstract}

\section{Introduction}
Single image super-resolution (SISR) aims at reconstructing a high-resolution (HR) image from a low-resolution (LR) one, which has proven useful in various tasks including object detection~\cite{bai2018sod}, face recognition~\cite{zou2011very, gunturk2003eigenface}, medical imaging~\cite{greenspan2008super}, and information forensics~\cite{lin20019forensics}. With the great success of deep learning, SRCNN~\cite{dong2015image} first introduces convolutional neural networks (CNNs) for SISR, outperforming classical approaches by large margins. After that, CNN-based SR methods focus on designing wider~\cite{lim2017enhanced,tong2017image,zhang2018residual} or deeper~\cite{haris2018deep,kim2016accurate,ledig2017photo,mao2016image,Zhang2017BeyondAG,zhang2017learning} network architectures for the performance gains. They require a high computational cost and a large amount of memory, and thus implementing them directly on a single chip for, \eg, televisions and mobile phones, is extremely hard without neural processing units and off-chip memory.     

\begin{figure}[t] % teaser image
  \captionsetup{font={small}}
  \centering
      \includegraphics[width=1\columnwidth]{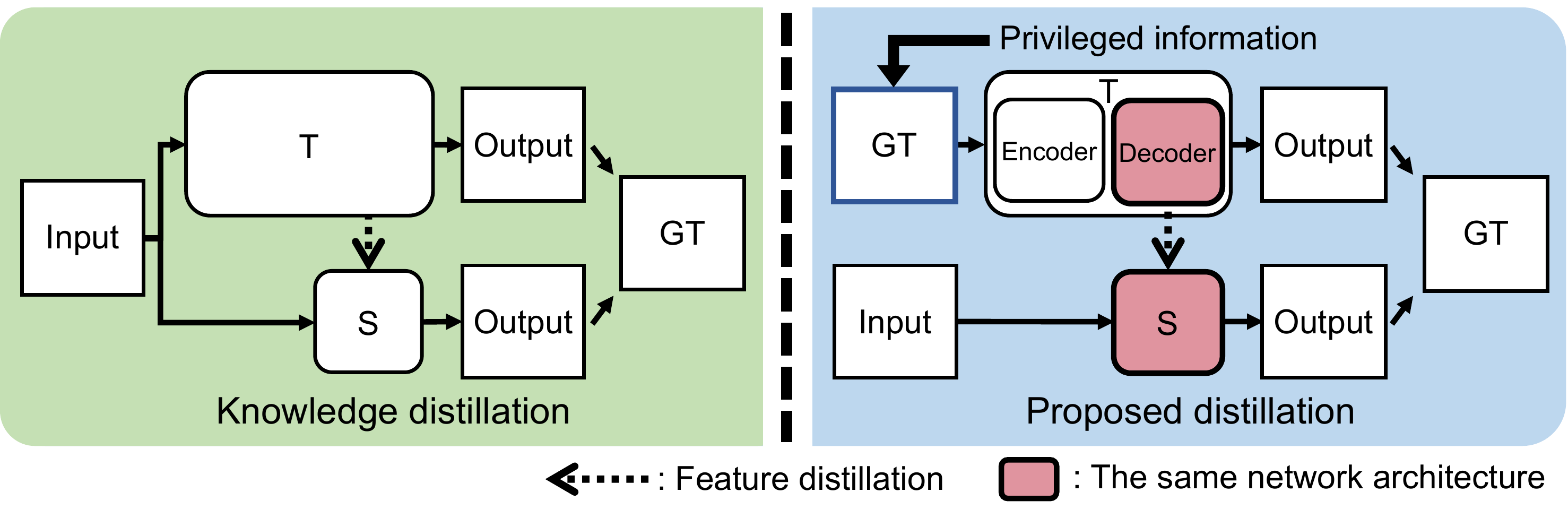}
      \caption{Compressing networks using knowledge distillation (left) transfers the knowledge from a large teacher model~(T) to a small student model~(S), with the same input,~\eg,~LR images in the case of SISR. Differently, the teacher in our framework (right) takes the ground truth~(\ie,~HR image) as an input, exploiting it as privileged information, and transfers the knowledge via feature distillation. (Best viewed in color.)
      }
  \label{fig:teaser}
\end{figure}

Many works introduce cost-effective network architectures~\cite{tai2017image,kim2016deeply,hui2018fast,ahn2018fast,han2016deep,han2015learning,dong2016accelerating} to reduce the computational burden and/or required memory, using recursive layers~\cite{tai2017image,kim2016deeply} or additional modules specific for SISR~\cite{hui2018fast,ahn2018fast}. Although they offer a good compromise in terms of PSNR and speed/memory, specially-designed or recursive architectures may be difficult to implement on hardware devices. Network pruning~\cite{han2015learning} and parameter quantization~\cite{han2016deep}, typically used for network compression, are alternative ways for efficient SR networks, where the pruning removes redundant connections of nodes and the quantization reduces bit-precision of weights or activations. The speedup achieved by the pruning is limited due to irregular memory accesses and poor data localizations~\cite{wen2016learning}, and the performance of the network quantization is inherently bound by that of a full-precision model. Knowledge distillation is another way of model compression, where a large model~(\ie,~a teacher network) transfers a softened version of the output distribution~(\ie,~logits)~\cite{hinton2014distilling} or intermediate feature representations~\cite{gao2018image,romero2015FitNets,ahn2019variational,heo2019comprehensive} to a small one~(\ie,~a student network), which has shown the effectiveness in particular for the task of image classification. Generalized distillation~\cite{lopez2016unifying} goes one step further, allowing a teacher to make use of extra~(privileged) information at training time, and assisting the training process of a student network with the complementary knowledge~\cite{hoffman2016learning,garcia2018modality}.

We present in this paper a simple yet effective framework for an efficient SISR method. The basic idea is that ground-truth HR images can be thought of as privileged information~(Fig.~\ref{fig:teaser}), which has not been explored in both SISR and privileged learning. It is true that the HR image includes the complementary information~(\eg,~high-frequency components) of LR images, but current SISR methods have used it just to penalize an incorrect reconstruction at the end of CNNs. On the contrary, our approach to using HR images as privileged information allows to extract the complementary features and leverage them explicitly for the SISR task. To implement this idea, we introduce a novel distillation framework where teacher and student networks try to reconstruct HR image but using different inputs (\ie,~ground-truth HR and corresponding LR images for the teacher and the student, respectively), which is clearly different from the conventional knowledge distillation framework (Fig.~\ref{fig:teaser}). Specifically, the teacher network has an hourglass architecture consisting of an encoder and a decoder. The encoder extracts compact features from HR images while encouraging them to imitate LR counterparts using an imitation loss. The decoder, which has the same network architecture as the student, reconstructs the HR images again using the compact features. Intermediate features in the decoder are then transferred to the student via feature distillation, such that the student learns the knowledge (\eg, high frequencies or fine details of HR inputs) of the teacher trained with the privileged data (\ie, HR image). Note that our framework is useful in that the student can be initialized with the network parameters of the decoder, which allows to transfer the reconstruction capability of the teacher to the student. We mainly exploit FSRCNN~\cite{dong2016accelerating} as the student network, since it has a hardware-friendly architecture~(\ie,~a stack of convolutional layers) and the number of parameters is extremely small compared to other CNN-based SR methods. Experimental results on standard SR benchmarks demonstrate the effectiveness of our approach, which boosts the performance of FSRCNN without any additional modules. To the best of our knowledge, our framework is the first attempt to leverage the privileged information for SISR. The main contributions of our work can be summarized as follows:

  \begin{itemize}[leftmargin=*]
    \item[$\bullet$] We present a novel distillation framework for SISR that leverages the ground truth~(\ie, HR images) as privileged information to transfer the important knowledge of the HR images to a student network.
    \item[$\bullet$] We propose to use an imitation loss to train a teacher network, making it possible to distill the knowledge a student is able to learn.
    \item[$\bullet$] We demonstrate that our approach boosts the performance of the current SISR methods, significantly, including FSRCNN~\cite{dong2016accelerating}, VDSR~\cite{kim2016accurate}, IDN~\cite{hui2018fast}, and CARN~\cite{ahn2018fast}. We show an extensive experimental analysis with ablation studies.
  \end{itemize} 

\section{Related work}
\subsubsection{SISR.} 
Early works on SISR design image priors to constrain the solution space~\cite{dai2009softedge,jian2008gradprior,yan2015gradsharp}, and leverage external datasets to learn the relationship between HR and LR images~\cite{freeman2002example,yang2008image,timo2013anchor,schulter2015fast,hong2004neighbor}, since lots of HR images can be reconstructed from a single LR image. CNNs have allowed remarkable advances in SISR. Dong~\emph{et al}. pioneer the idea of exploiting CNNs for SISR, and propose SRCNN~\cite{dong2015image} that learns a mapping function directly from input LR to output HR images. Recent methods using CNNs exploit a much larger number of convolutional layers. Sparse~\cite{lim2017enhanced,ledig2017photo,mao2016image} or dense~\cite{tong2017image,zhang2018residual,haris2018deep} skip connections between them prevent a gradient vanishing problem, achieving significant performance gains over classical approaches. More recently, efficient networks for SISR in terms of memory and/or runtime have been introduced. Memory-efficient SR methods~\cite{kim2016deeply,tai2017image,tai2017memnet,li2019feedback} reduce the number of network parameters by reusing them recursively. They further improve the reconstruction performance using residual units~\cite{tai2017image}, memory~\cite{tai2017memnet} or feedback~\cite{li2019feedback} modules but at the cost of runtime. Runtime-efficient methods~\cite{dong2016accelerating,hui2018fast,ahn2018fast,hui2019lightweight} on the other hand are computationally cheap. They use cascaded~\cite{ahn2018fast} or multi-branch~\cite{hui2018fast,hui2019lightweight} architectures, or exploit group convolutions~\cite{xie2017aggre,chollet2017xcep}. The main drawback of such SR methods is that their hardware implementations are difficult due to the network architectures specially-designed for the SR task. FSRCNN~\cite{dong2016accelerating} reduces both runtime and memory. It uses typical convolutional operators with a small number of filters and feature channels, except the deconvolution layer at the last part of the network. Although FSRCNN has a hardware-friendly network architecture, it is largely outperformed by current SR methods.

\subsubsection{Feature distillation.}
The purpose of knowledge distillation is to transfer the representation ability of a large model~(teacher) to a small one~(student) for enhancing the performance of the student model. It has been widely used to compress networks, typically for classification tasks. In this framework, the softmax outputs of a teacher are regarded as soft labels, providing informative clues beyond discrete labels~\cite{hinton2014distilling}. Recent methods extend this idea to feature distillation, which transfers intermediate feature maps~\cite{romero2015FitNets,ahn2019variational}, their transformations~\cite{heo2019comprehensive,Zagoruyko2017AT}, the differences of features before and after a stack of layers~\cite{yim2017gift}, or pairwise relations within feature maps~\cite{liu2019structured}. In particular, the variational information distillation~(VID) method~\cite{ahn2019variational} transfers the knowledge by maximizing the mutual information between feature maps of teacher and student networks. We exploit VID for feature distillation, but within a different framework. Instead of sharing the same inputs~(\ie, LR images) with the student, our teacher network inputs HR images, that contain the complementary information of LR images, to take advantage of privileged information.

Closely related to ours, SRKD~\cite{gao2018image} applies the feature distillation technique to SISR in order to compress the size of SR network, where a student is trained to have similar feature distributions to those of a teacher. Following the conventional knowledge distillation, the student and teacher networks in SRKD use the same inputs of LR images. This is clearly different from our method in that our teacher takes ground-truth HR images as inputs, allowing to extract more powerful feature representations for image reconstruction.

\subsubsection{Generalized distillation.}
Learning using privileged information~\cite{vapnik2009new,vapnik2015learning} is a machine learning paradigm that uses extra information, which requires an additional cost, at training time, but with no accessibility to it at test time. In a broader context, generalized distillation~\cite{lopez2016unifying} covers both feature distillation and learning using privileged information. The generalized distillation enables transferring the privileged knowledge of a teacher to a student. For example, the works of~\cite{hoffman2016learning,garcia2018modality} adopt the generalized distillation approach for object detection and action recognition, where depth images are used as privileged information. In the framework, a teacher is trained to extract useful features from depth images. They are then transferred to a student which takes RGB images as inputs, allowing the student to learn complementary representations from privileged information. Our method belongs to generalized distillation, since we train a teacher network with ground-truth HR images, which can be viewed as privileged information, and transfer the knowledge to a student network. Different from previous methods, our method does not require an additional cost for privileged information, since the ground truth is readily available at training time.

\begin{figure}[t] % Overview figure
  \captionsetup{font={small}}
  \centering
  \includegraphics[width=1\textwidth]{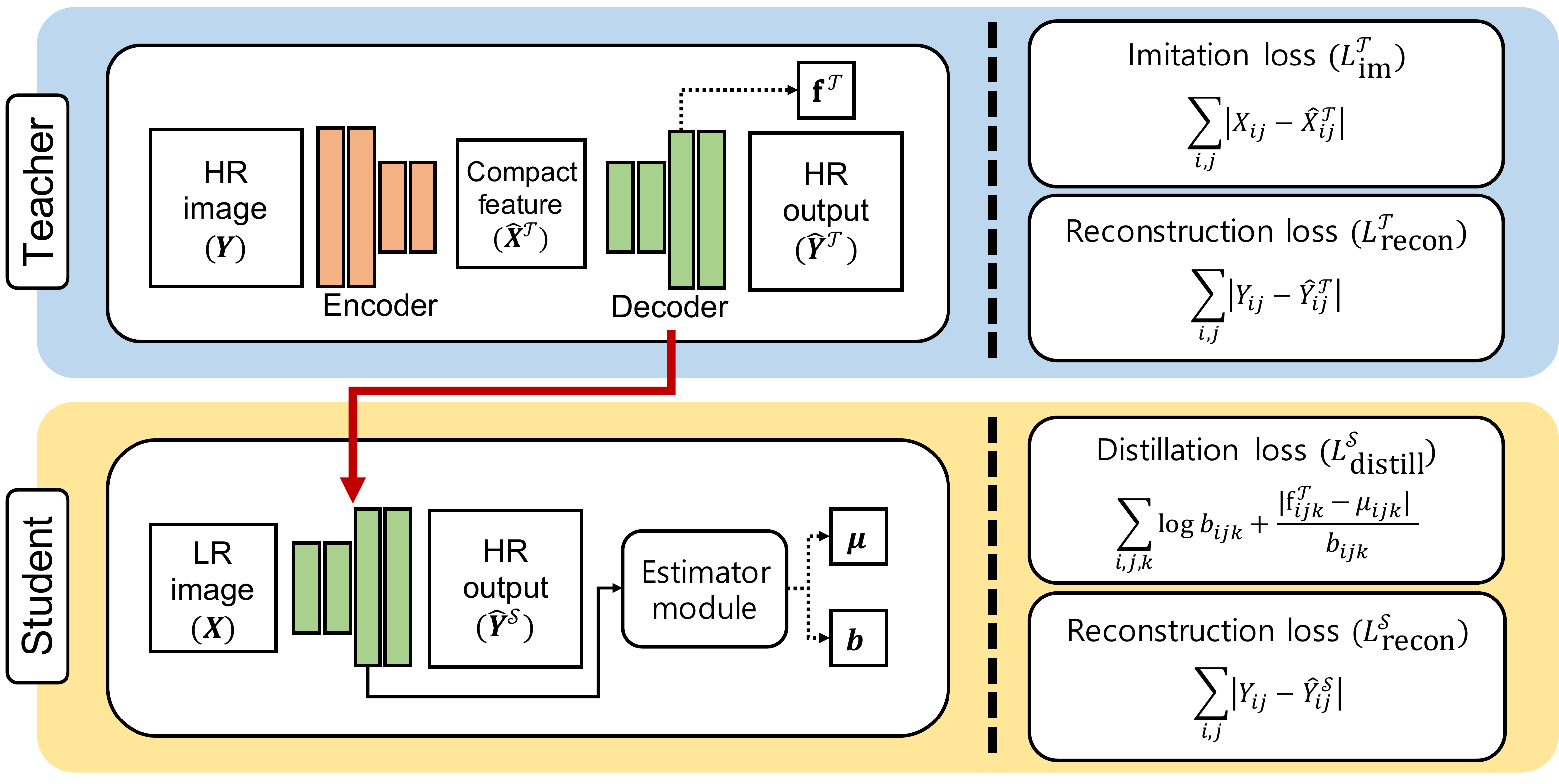} 
  \caption{{{\bf{Overview of our framework.}} A teacher network inputs a HR image~$\mathbf{Y}$ and extracts a compact feature representation~$\hat{\mathbf{X}}^{\mathcal{T}}$ using an encoder. The decoder in the network then reconstructs a HR output~$\hat{\mathbf{Y}}^{\mathcal{T}}$. To train the teacher network, we use imitation~$L^{\mathcal{T}}_{\textrm{im}}$ and reconstruction~$L^{\mathcal{T}}_{\textrm{recon}}$ losses. After training the teacher, a student network is initialized with weights of the decoder in the teacher network~(red line), and restores a HR output~$\hat{\mathbf{Y}}^{\mathcal{S}}$ from a LR image~$\mathbf{X}$. Note that the student network and the decoder share the same network architecture. The estimator module takes intermediate feature maps of the student network, and outputs location and scale maps, $\bm{\mu}$ and $\mathbf{b}$, respectively. To train the student network, we exploit a reconstruction loss~$L^{\mathcal{S}}_{\textrm{recon}}$ together with a distillation loss~$L^{\mathcal{S}}_{\textrm{distill}}$ using the intermediate representation~$\mathbf{f}^{\mathcal{T}}$ of the teacher network and the parameter maps of $\bm{\mu}$ and $\mathbf{b}$. See text for details. (Best viewed in color.)
  }}
  \label{fig:overview}
\end{figure}

\section{Method}
We denote by $\mathbf{X}$ and $\mathbf{Y}$ LR and ground-truth HR images. Given the LR image~$\mathbf{X}$, we reconstruct a high-quality HR output~$\hat{\mathbf{Y}}^{\mathcal{S}}$ efficiently in terms of both speed and memory. To this end, we present an effective framework consisting of teacher and student networks. The teacher network learns to distill the knowledge from privileged information (\ie,~a ground-truth HR image~$\mathbf{Y}$). After training the teacher network, we transfer the knowledge distilled from the teacher to the student to boost the reconstruction performance. We show in Fig.~\ref{fig:overview} an overview of our framework.

\subsection{Teacher}\label{sec:teacher}
In order to transfer knowledge from a teacher to a student, the teacher should be superior to the student, while extracting informative features. To this end, we treat ground-truth HR images as privileged information, and exploit an \textit{intelligent teacher}~\cite{vapnik2015learning}.  As will be seen in our experiments, the network architecture of the teacher influences the SR performance significantly. As the teacher network inputs ground-truth HR images, it may not be able to extract useful features, and just learn to copy the inputs for the reconstruction of HR images, regardless of its capacity. Moreover, a large difference for the number of network parameters or the performance gap between the teacher and the student discourages the distillation process~\cite{cho2019efficacy,mirzadeh2019improved}. To reduce the gap while promoting the teacher to capture useful features, we exploit an hourglass architecture for the teacher network. It projects the HR images into a low-dimensional feature space to generate compact features, and reconstructs the original HR images from them, such that the teacher learns to extract better feature representations for an image reconstruction task. Specifically, the teacher network consists of an encoder~$G^{\mathcal{T}}$ and a decoder~$F^{\mathcal{T}}$. Given a pair of LR and HR images, the encoder~$G^{\mathcal{T}}$ transforms the input HR image~$\mathbf{Y}$ into the feature representation~$\hat{\mathbf{X}}^{\mathcal{T}}$ in a low-dimensional space: 
  \begin{equation}
    \hat{\mathbf{X}}^{\mathcal{T}} = G^{\mathcal{T}}(\mathbf{Y}),
  \end{equation}
where the feature representation of~$\hat{\mathbf{X}}^{\mathcal{T}}$ has the same size as the LR image. The decoder~$F^{\mathcal{T}}$ reconstructs the HR image~$\hat{\mathbf{Y}}^{\mathcal{T}}$ using the compact feature~$\hat{\mathbf{X}}^{\mathcal{T}}$:
  \begin{equation}
    \hat{\mathbf{Y}}^{\mathcal{T}} = F^{\mathcal{T}}(\hat{\mathbf{X}}^{\mathcal{T}}).
  \end{equation}
For the decoder, we use the same architecture as the student network. It allows the teacher to have a similar representational capacity as the student, which has proven useful in~\cite{mirzadeh2019improved}.

\subsubsection{Loss.}
To train the teacher network, we use reconstruction and imitation losses, denoted by~$L^{\mathcal{T}}_{\textrm{recon}}$ and~$L^{\mathcal{T}}_{\textrm{im}}$, respectively. The reconstruction term computes the mean absolute error~(MAE) between the HR image~$\mathbf{Y}$ and its reconstruction~$\hat{\mathbf{Y}}^{\mathcal{T}}$ defined as: 
   \begin{equation}
      L^{\mathcal{T}}_{\textrm{recon}} = \frac{1}{HW} \sum^{H}_{i=1} \sum^{W}_{j=1} \vert Y_{ij} - \hat{Y}^{\mathcal{T}}_{ij} \vert,
        \label{eq:recon_t}
   \end{equation}
where $H$ and $W$ are height and width of the HR image, respectively, and we denote by $Y_{ij}$ an intensity value of $\mathbf{Y}$ at position~$(i,j)$. It encourages the encoder output~(\ie, compact feature $\hat{\mathbf{X}}^{\mathcal{T}}$) to contain useful information for the image reconstruction and forces the decoder to reconstruct the HR image again using the compact feature. The imitation term restricts the representational power of the encoder, making the output of the encoder close to the LR image. Concretely, we define this term as the MAE between the LR image~$\mathbf{X}$ and the encoder output~$\hat{\mathbf{X}}^{\mathcal{T}}$:
   \begin{equation}
      L^{\mathcal{T}}_{\textrm{im}}  = \frac{1}{H^{\prime}W^{\prime}} \sum_{i=1}^{H^{\prime}} \sum_{j=1}^{W^{\prime}} \vert X_{ij} - \hat{X}^{\mathcal{T}}_{ij} \vert,
   \end{equation}
where $H^{\prime}$ and $W^{\prime}$ are height and width of the LR image, respectively. This facilitates an initialization of the student network that takes the LR image~$\mathbf{X}$ as an input. Note that our framework avoids the trivial solution that the compact feature becomes the LR image since the network parameters in the encoder are updated by both the imitation and reconstruction terms. The overall objective is a sum of reconstruction and imitation terms, balanced by the parameter~$\lambda^{\mathcal{T}}$: 
   \begin{equation}
      L^{\mathcal{T}}_{\textrm{total}} = L^{\mathcal{T}}_{\textrm{recon}} + \lambda^{\mathcal{T}} L^{\mathcal{T}}_{\textrm{im}}.
   		\label{eq:all}
   \end{equation}

\subsection{Student}\label{sec:student}
A student network has the same architecture as the decoder~$F^{\mathcal{T}}$ in the teacher, but uses a different input. It takes a LR image~$\mathbf{X}$ as an input and generates a HR image~$\hat{\mathbf{Y}}^{\mathcal{S}}$:
  \begin{equation}
    \hat{\mathbf{Y}}^{\mathcal{S}} = F^{\mathcal{S}}(\mathbf{X}).
  \end{equation}
We initialize the weights of the student network with those of the decoder in the teacher. This transfers the reconstruction capability of the teacher to the student and provides a good starting point for optimization. Note that several works~\cite{hoffman2016learning,garcia2018modality} point out that how to initialize network weights is crucial for the performance of a student. We adopt FSRCNN~\cite{dong2016accelerating}, a hardware-friendly SR architecture, as the student network~$F^{\mathcal{S}}$.

\subsubsection{Loss.}
Although the network parameters of the student~$F^{\mathcal{S}}$ and the decoder~$F^{\mathcal{T}}$ in the teacher are initially set to the same, the features extracted from them are different due to the different inputs. Besides, these parameters are not optimized with input LR images. We further train the student network~$F^{\mathcal{S}}$ with a reconstruction loss~$L^{\mathcal{S}}_{\textrm{recon}}$ and a distillation loss~$L^{\mathcal{S}}_{\textrm{distill}}$. The reconstruction term is similarly defined as Eq.~\eqref{eq:recon_t} using the ground-truth HR image and its reconstruction from the student network, dedicating to the SISR task:
  \begin{equation}
    L^{\mathcal{S}}_{\textrm{recon}} = \frac{1}{HW} \sum^{H}_{i=1} \sum^{W}_{j=1} \vert Y_{ij} - \hat{Y}^{\mathcal{S}}_{ij} \vert.
  \end{equation}              
The distillation term focuses on transferring the knowledge of the teacher to the student. Overall, we use the following loss to train the student network: 
  \begin{equation}
    L^{\mathcal{S}}_{\textrm{total}} = L^{\mathcal{S}}_{\textrm{recon}} + \lambda^{\mathcal{S}} L^{\mathcal{S}}_{\textrm{distill}},
  \end{equation}
where $\lambda^{\mathcal{S}}$ is a distillation parameter. In the following, we describe the distillation loss in detail.

We adopt the distillation loss proposed in the VID method~\cite{ahn2019variational}, which maximizes mutual information between the teacher and the student. We denote by~$\mathbf{f}^{\mathcal{T}}$ and~$\mathbf{f}^{\mathcal{S}}$ the intermediate feature maps of the teacher and student networks, respectively, having the same size of $C \times H^{\prime} \times W^{\prime}$, where $C$ is the number of channels. We define mutual information~$I(\mathbf{f}^{\mathcal{T}};\mathbf{f}^{\mathcal{S}})$ as follows:
  \begin{equation}
    I(\mathbf{f}^{\mathcal{T}};\mathbf{f}^{\mathcal{S}}) = H(\mathbf{f}^{\mathcal{T}}) - H(\mathbf{f}^{\mathcal{T}} \vert \mathbf{f}^{\mathcal{S}}),
  \end{equation}
where we denote by~$H(\mathbf{f}^{\mathcal{T}})$ and $H(\mathbf{f}^{\mathcal{T}} \vert \mathbf{f}^{\mathcal{S}})$ marginal and conditional entropies, respectively. To maximize the mutual information, we should minimize the conditional entropy~$H(\mathbf{f}^{\mathcal{T}} \vert \mathbf{f}^{\mathcal{S}})$. However, an exact optimization w.r.t the weights of the student is intractable, as it involves an integration over a conditional probability~$p(\mathbf{f}^{\mathcal{T}} \vert \mathbf{f}^{\mathcal{S}})$. The variational information maximization technique~\cite{barber2003algorithm} instead approximates the conditional distribution~$p(\mathbf{f}^{\mathcal{T}} \vert \mathbf{f}^{\mathcal{S}})$ using a parametric model~$q(\mathbf{f}^{\mathcal{T}} \vert \mathbf{f}^{\mathcal{S}})$, such as the Gaussian or Laplace distributions, making it possible to find a lower bound of the mutual information~$I(\mathbf{f}^{\mathcal{T}};\mathbf{f}^{\mathcal{S}})$. Using this technique, we maximize the lower bound of mutual information~$I(\mathbf{f}^{\mathcal{T}};\mathbf{f}^{\mathcal{S}})$ for feature distillation. As the parametric model~$q(\mathbf{f}^{\mathcal{T}} | \mathbf{f}^\mathcal{S})$, we use a multivariate Laplace distribution with parameters of location and scale, $\bm{\mu} \in \mathds{R} ^{C \times H^\prime \times W^\prime}$ and $\mathbf{b} \in \mathds{R}^{C \times H^\prime \times W^\prime}$, respectively. We define the distillation loss~$L^{\mathcal{S}}_{\textrm{distill}}$ as follows:
  \begin{equation}
    L^{\mathcal{S}}_{\textrm{distill}} = \frac{1}{C H^{\prime} W^{\prime}} \sum_{i=1}^{C} \sum_{j=1}^{H^{\prime}} \sum_{k=1}^{W^{\prime}} \log{b_{ijk}} 
    + \frac{\vert f^{\mathcal{T}}_{ijk} - \mu_{ijk} \vert}{b_{ijk}},
    \label{eq:distill}
  \end{equation}
where we denote by $\mu_{ijk}$ the element of $\bm{\mu}$ at the position~$(i,j,k)$. This minimizes the distance between the features~$\mathbf{f}^{\mathcal{T}}$ of the teacher and the location map~$\bm{\mu}$. The scale map~$\mathbf{b}$ controls the extent of distillation. For example, when the student does not benefit from the distillation, the scale parameter~$b_{ijk}$ increases in order to reduce the extent of distillation. This is useful for our framework where the teacher and student networks take different inputs, since it adaptively determines the features the student is affordable to learn from the teacher. The term~$\log{b_{ijk}}$ prevents a trivial solution where the scale parameter goes to infinite. We estimate these maps of $\bm{\mu}$ and $\mathbf{b}$ from the features of the student~$\mathbf{f}^{\mathcal{S}}$. Note that other losses designed for feature distillation can also be used in our framework (See the supplementary material).

\subsubsection{Estimator module.}
We use a small network to estimate the parameters of location~$\bm{\mu}$ and scale~$\mathbf{b}$ in Eq.~\eqref{eq:distill}. It consists of location and scale branches, where each takes the features of the student~$\mathbf{f}^{\mathcal{S}}$ and estimates the location and scale maps, separately. Both branches share the same network architecture of two $1\times1$ convolutional layers and a PReLU~\cite{he2015delving} between them. For the scale branch, we add the softplus function~$(\zeta(x) = \log(1 + e^x))$~\cite{dugas2001incorporating} at the last layer, forcing the scale parameter to be positive. Note that the estimation module is used only at training time.

\section{Experiments}

\subsection{Experimental details}\label{sec:detail}
\subsubsection{Implementation details.} The encoder in the teacher network consists of 4 blocks of convolutional layers followed by a PReLU~\cite{he2015delving}. All the layers, except the second one, perform convolutions with stride 1. In the second block, we use the convolution with stride~$s$~(\ie,~a scale factor) to downsample the size of the HR image to that of the LR image. The kernel sizes of the first two and the last two blocks are 5 $\times$ 5 and 3 $\times$ 3, respectively. The decoder in the teacher and the student network have the same architecture as FSRCNN~\cite{dong2016accelerating} consisting of five components: Feature extraction, shrinking, mapping, expanding, and deconvolution modules. We add the estimator module for location and scale maps on top of the expanding module in the student network. We use these maps together with the output features of the expanding module in the decoder to compute the distillation loss. We set the hyperparameters for losses using a grid search on the DIV2K dataset~\cite{timofte2017ntire}, and choose the ones~($\lambda^{\mathcal{T}}=10^{-4}$,~$\lambda^{\mathcal{S}}=10^{-6}$) that give the best performance. We implement our framework using 
\texttt{PyTorch}~\cite{paszke2017automatic}.

\subsubsection{Training.} To train our network, we use the training split of DIV2K~\cite{timofte2017ntire} corresponding 800 pairs of LR and HR images, where the LR images are synthesized by bicubic downsampling. We randomly crop HR patches of size 192 $\times$ 192 from the HR images. LR patches are cropped from the corresponding LR images according to the scale factor. For example, LR patches of size 96 $\times$ 96 are used for the scale factor of 2. We use data augmentation techniques, including random rotation and horizontal flipping. The teacher network is trained with random initialization. We train our model with a batch size of 16 about 1000k iterations over the training data. We use the Adam~\cite{kingma2014adam} with $\beta_1=0.9$ and $\beta_2=0.999$. As a learning rate, we use $10^{-3}$ and reduce it until $10^{-5}$ using a cosine annealing technique~\cite{loshchilov2016sgdr}.

\subsubsection{Evaluation.} We evaluate our framework on standard benchmarks including Set5~\cite{bevilacqua2012low}, Set14~\cite{zeyde2010single}, B100~\cite{martin2001database}, and Urban100~\cite{huang2015single}. Following the experimental protocol in~\cite{lim2017enhanced}, we use the peak signal to noise ratio (PSNR) and the structural similarity index (SSIM)~\cite{wang2004image} on the luminance channel as evaluation metrics. 

\begin{table}[t]
  \setlength{\tabcolsep}{0.3em}

  \captionsetup{font={small}}
  \caption{Average PSNR of student and teacher networks, trained with variants of our framework, on the Set5~\cite{bevilacqua2012low} dataset. We use FSRCNN~\cite{dong2016accelerating}, reproduced by ourselves using the DIV2K~\cite{timofte2017ntire} dataset without distillation, as the baseline in the first row. We denote by VID$_{G}$ and VID$_{L}$ VID losses~\cite{ahn2019variational} with the Gaussian and Laplace distributions, respectively. The performance gains of each component over the baseline are shown in the parentheses. The number in bold indicates the best performance and underscored one is the second best.} 
  \begin{adjustbox}{width=0.91\columnwidth,center}
  \centering
    \begin{tabular}{C{2.0cm} C{1.5cm} C{1.5cm} C{1.5cm} | C{2.5cm} C{1.5cm}}
      \multirow{2}{*}{\shortstack[1]{Hourglass \\ architecture}} & 
      \multirow{2}{*}{\shortstack{Weight \\ transfer}} &  
      \multirow{2}{*}{\shortstack[1]{$L^{\mathcal{T}}_{\textrm{im}}$}} &  
      \multirow{2}{*}{\shortstack{$L^{\mathcal{S}}_{\textrm{distill}}$}} & 
      \multicolumn{1}{c}{\multirow{2}{*}{\shortstack{Student \\ PSNR}}} &
      \multicolumn{1}{c}{\multirow{2}{*}{\shortstack{Teacher \\ PSNR}}} \\
      &&&&&\\
      \hline
      \hline
      &&&&&\\[-0.9em]
      -        &  -        &  -        &  -        &  37.15~(baseline)&  - \\
      \xmark   &  -        &  -        &  MAE       &  37.19~(+0.04)  & 57.60 \\
      \cmark   &  \xmark   &  \xmark   &  MAE       &  37.22~(+0.07)  & 37.70 \\
      \cmark   &  \cmark   &  \xmark   &  MAE       &  37.23~(+0.08)  & 37.70 \\
      \cmark   &  \cmark   &  \cmark   &  MAE       &  37.27~(+0.12)  & 37.65 \\
      \cmark   &  \cmark   &  \cmark   &  VID$_{G}$~\cite{ahn2019variational}      &  \underline{37.31}~(+0.16) & 37.65  \\
      \cmark   &  \cmark   &  \cmark   &  VID$_{L}$~\cite{ahn2019variational}      &  {\bf{37.33}}~(+0.18) & 37.65 \\
      \hline
    \end{tabular}
    \label{tab:ablation_components}
    \end{adjustbox}
\end{table}

\subsection{Ablation studies}\label{sec:ablation}
We present an ablation analysis on each component of our framework. We report quantitative results in terms of the average PSNR on Set5~\cite{bevilacqua2012low} with the scale factor of 2. The results on a large dataset (\ie, B100~\cite{martin2001database}) can be seen in the supplementary material. We show in Table~\ref{tab:ablation_components} the average PSNR for student networks trained with variants of our framework. The results of the baseline in the first row are obtained using FSRCNN~\cite{dong2016accelerating}. From the second row, we can clearly see that feature distillation boosts the PSNR performance. For the teacher network in the second row, we use the same network architecture as FSRCNN except for the deconvolution layers. In contrast to FSRCNN, the teacher inputs HR images, and thus we replace the deconvolution layer with a convolutional layer, preserving the size of the inputs. We can see from the third row that a teacher network with an hourglass architecture improves the student performance. The hourglass architecture limits the performance of the teacher and degrades the performance (\eg, a 19.9dB decrease compared to that of the teacher in the second row), reducing the performance gap between the teacher and the student. This allows the feature distillation to be more effective, thus the student of the third row performs better (37.22dB) than that of the second row (37.19dB), which can also be found in recent works~\cite{cho2019efficacy,mirzadeh2019improved}. The fourth row shows that the student network benefits from initializing the network weights with those of the decoder in the teacher, since this provides a good starting point for learning, and transfers the reconstruction capability of the teacher. From the fifth row, we observe that an imitation loss further improves the PSNR performance, making it easier for the student to learn features from the teacher. The next two rows show that the VID loss~\cite{ahn2019variational}, especially with the Laplace distribution~(VID$_L$), provides better results than the MAE, and combining all components gives the best performance. The distillation loss based on the MAE forces the feature maps of the student and teacher networks to be the same. This strong constraint on the feature maps is, however, problematic in our case, since we use different inputs for the student and teacher networks. The VID method allows the student to learn important features adaptively. We also compare the performance of our framework and a typical distillation approach with different losses in the supplementary material.

\begin{figure}[t]
    \captionsetup{font={small}}
    \captionsetup[subfigure]{aboveskip=1pt,belowskip=-0.5pt}
    \centering
    \begin{adjustbox}{width=0.77\columnwidth,center}
    \begin{subfigure}{0.2\textwidth}
        \caption*{\scriptsize HR}
        \includegraphics[width=\textwidth]{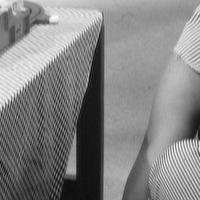}
    \end{subfigure}
    \begin{subfigure}{0.2\textwidth}
        \caption*{\scriptsize LR}
        \includegraphics[width=\textwidth]{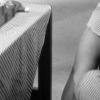}
    \end{subfigure}
    \begin{subfigure}{0.2\textwidth}
        \caption*{\scriptsize $\hat{X}^{\mathcal{T}}$ w/ $L^{\mathcal{T}}_{\rm{im}}$}
        \includegraphics[width=\textwidth]{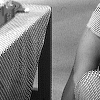}
    \end{subfigure}
    \begin{subfigure}{0.2\textwidth}
        \caption*{\scriptsize $\hat{X}^{\mathcal{T}}$ w/o $L^{\mathcal{T}}_{\rm{im}}$}
        \includegraphics[width=\textwidth]{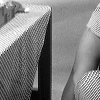}
    \end{subfigure}
    \end{adjustbox}
    
    \begin{adjustbox}{width=0.77\columnwidth,center}    
    \begin{subfigure}{0.354\textwidth} % 3x set14 12th
        \includegraphics[width=\textwidth]{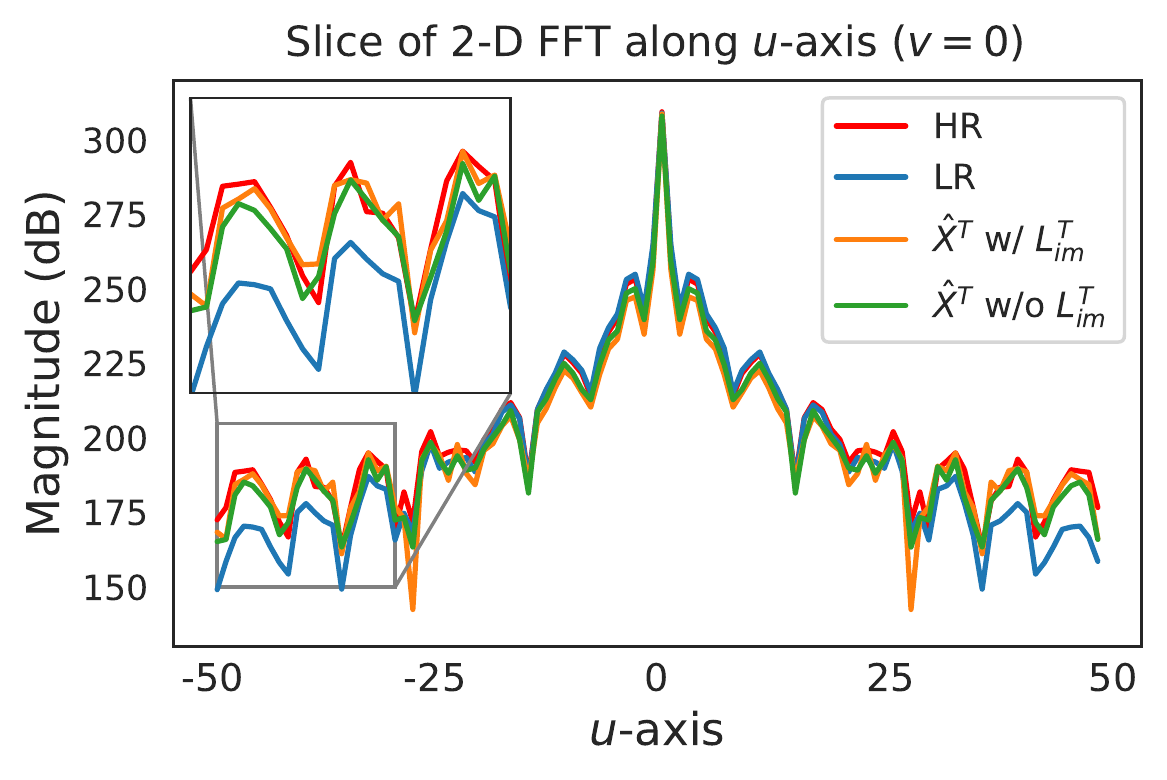}
    \end{subfigure}
    \begin{subfigure}{0.224\textwidth}
        \includegraphics[width=\textwidth]{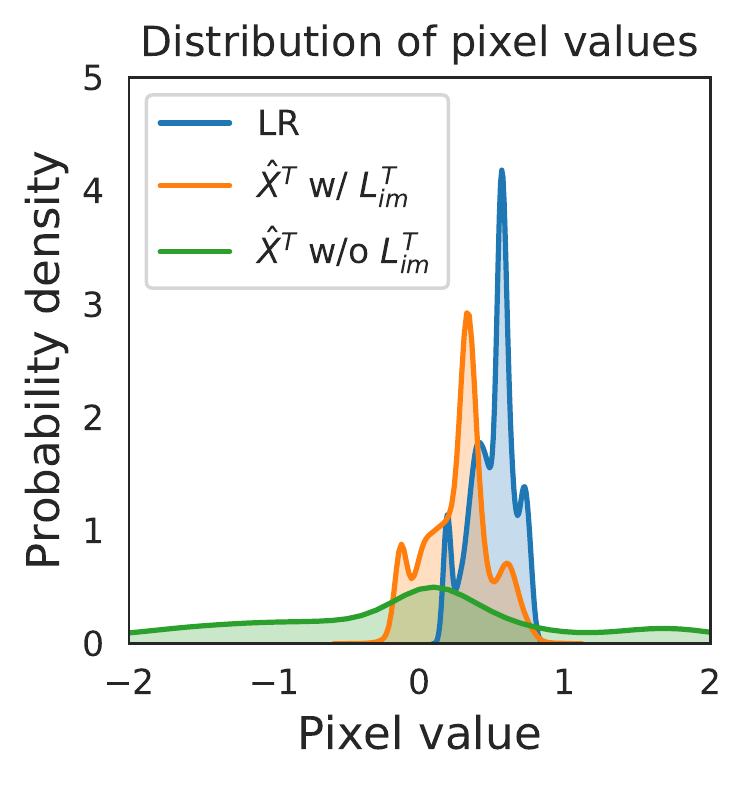}
    \end{subfigure}
    \end{adjustbox}
    \vfill
    \caption{Analysis on compact features in spatial (top) and frequency (bottom left) domains and the distribution of pixel values (bottom right). To visualize the compact features in the frequency domain, we apply the 2D Fast Fourier Transform (FFT) to the image, obtaining its magnitude spectrum. It is then sliced along the $u$-axis. (Best viewed in color.)}
    \label{fig:compact_feature}
\end{figure}

\subsection{Analysis on compact features}\label{sec:compact_feature}
In Fig. \ref{fig:compact_feature}, we show an analysis on compact features in spatial and frequency domains. Compared to the LR image, the compact features~$\hat{\mathbf{X}}^{\mathcal{T}}$ show high-frequency details regardless of whether the imitation loss~$L^{\mathcal{T}}_{\rm{im}}$ is used or not. This can also be observed in the frequency domain -- The compact features contain more high-frequency components than the LR image, and the magnitude spectrums of them are more similar to that of the HR image especially for high-frequency components. By taking these features as inputs, the decoder in the teacher shows the better performance than the student (Table \ref{tab:ablation_components}) despite the fact that they have the same architecture. This demonstrates that the compact features extracted from the ground truth contain useful information for reconstructing the HR image, encouraging the student to reconstruct more accurate results via feature distillation. In the bottom right of the Fig. \ref{fig:compact_feature}, we can see that the pixel distributions of the LR image and the compact feature are largely different without the imitation loss, discouraging the weight transfer to the student. The imitation loss~$L^{\mathcal{T}}_{\rm{im}}$ alleviates this problem by encouraging the distributions of the LR image and the compact feature to be similar. 

\begin{table}[t]
    \setlength{\tabcolsep}{0.3em}
    \captionsetup{font={small}}
    \caption{Quantitative comparison with the state of the art on SISR. We report the average PSNR/SSIM for different scale factors~(2$\times$, 3$\times$, and 4$\times$) on Set5~\cite{bevilacqua2012low}, Set14~\cite{zeyde2010single}, B100~\cite{martin2001database}, and Urban100~\cite{huang2015single}. *: models reproduced by ourselves using the DIV2K~\cite{timofte2017ntire} dataset without distillation; Ours: student networks of our framework.}
    \begin{adjustbox}{width=\columnwidth,center}
      \small
      \begin{tabular}{c|l|c|c|c|C{2.1cm}|C{2.1cm}|C{2.1cm}|C{2.1cm}}
        \hline
        \multicolumn{1}{c|}{\multirow{2}{*}{Scale}} & \multicolumn{1}{c|}{\multirow{2}{*}{Methods}} & \multicolumn{1}{c|}{\multirow{2}{*}{Param.}} & 
        \multicolumn{1}{c|}{\multirow{2}{*}{MultiAdds}} & \multicolumn{1}{c|}{\multirow{2}{*}{Runtime}} & Set5~\cite{bevilacqua2012low} & Set14~\cite{zeyde2010single} & B100~\cite{martin2001database} & Urban100~\cite{huang2015single} \\
        & & & & & PSNR/SSIM & PSNR/SSIM & PSNR/SSIM & PSNR/SSIM \\ 
        \hline
        \hline
        \multirow{11}{*}{2}
        & FSRCNN~\cite{dong2016accelerating} & 13K & 6.0G & 0.83ms & 37.05/0.9560 & 32.66/0.9090 & 31.53/\underline{0.8920} & 29.88/0.9020 \\
        & FSRCNN* & 13K & 6.0G & 0.83ms & \underline{37.15}/\underline{0.9568} & \underline{32.71}/\underline{0.9095} & \underline{31.58}/0.8913 & \underline{30.05}/\underline{0.9041} \\
        & FSRCNN~(Ours) & 13K & 6.0G & 0.83ms & {\bf{37.33}}/{\bf{0.9576}} & {\bf{32.79}}/{\bf{0.9105}} & {\bf{31.65}}/{\bf{0.8926}} & {\bf{30.24}}/{\bf{0.9071}} \\
        \cline{2-9}
        & Bicubic Int.& - & - & - & 33.66/0.9299 & 30.24/0.8688 & 29.56/0.8431 & 26.88/0.8403 \\
        & DRCN~\cite{kim2016deeply} & 1,774K & 17,974.3G & 239.93ms & 37.63/0.9588 & 33.04/0.9118 & 31.85/0.8942 & 30.75/0.9133 \\
        & DRRN~\cite{tai2017image} & 297K & 6,796.9G & 105.76ms & 37.74/0.9591 & 33.23/0.9136 & 32.05/0.8973 & 31.23/0.9188 \\
        & MemNet~\cite{tai2017memnet} & 677K & 2,662.4G & 21.06ms & 37.78/0.9597 & 33.28/0.9142 & 32.08/0.8978 & 31.31/0.9195 \\
        & CARN~\cite{ahn2018fast} & 1,592K & 222.8G & 8.43ms & 37.76/0.9590 & 33.52/0.9166 & 32.09/0.8978 & 31.92/0.9256 \\
        & IDN~\cite{hui2018fast} & 591K & 136.5G & 7.01ms & 37.83/0.9600 & 33.30/0.9148 & 32.08/0.8985 & 31.27/0.9196 \\
        & SRFBN~\cite{li2019feedback} & 3,631K & 1,126.7G & 108.52ms & 38.11/0.9609 & 33.82/0.9196 & 32.29/0.9010 & 32.62/0.9328 \\
        & IMDN~\cite{hui2019lightweight} & 694K & 159.6G & 6.97ms & 38.00/0.9605 & 33.63/0.9177 & 32.19/0.8996 & 32.17/0.9283 \\
        \hline
        \multirow{11}{*}{3}
        & FSRCNN~\cite{dong2016accelerating} & 13K & 5.0G & 0.72ms & \underline{33.18}/0.9140 & 29.37/0.8240 & \underline{28.53}/\underline{0.7910} & 26.43/0.8080 \\
        & FSRCNN* & 13K & 5.0G & 0.72ms & 33.15/\underline{0.9157} & \underline{29.45}/\underline{0.8250} & 28.52/0.7895 & \underline{26.49}/\underline{0.8089} \\
        & FSRCNN~(Ours) & 13K & 5.0G & 0.72ms & {\bf{33.31}}/{\bf{0.9179}} & {\bf{29.57}}/{\bf{0.8276}} & {\bf{28.61}}/{\bf{0.7919}} & {\bf{26.67}}/{\bf{0.8153}} \\
        \cline{2-9}
        & Bicubic Int.& - & - & - & 30.39/0.8682 & 27.55/0.7742 & 27.21/0.7385 & 24.46/0.7349 \\
        & DRCN~\cite{kim2016deeply} & 1,774K & 17,974.3G & 239.19ms & 33.82/0.9226 & 29.76/0.8311 & 28.80/0.7963 & 27.15/0.8276 \\
        & DRRN~\cite{tai2017image} & 297K & 6,796.9G & 98.58ms & 34.03/0.9244 & 29.96/0.8349 & 28.95/0.8004 & 27.53/0.8378 \\
        & MemNet~\cite{tai2017memnet} & 677K & 2,662.4G & 11.33ms & 34.09/0.9248 & 30.00/0.8350 & 28.96/0.8001 & 27.56/0.8376 \\
        & CARN~\cite{ahn2018fast} & 1,592K & 118.8G & 3.86ms & 34.29/0.9255 & 30.29/0.8407 & 29.06/0.8034 & 28.06/0.8493 \\
        & IDN~\cite{hui2018fast} & 591K & 60.6G & 3.62ms & 34.11/0.9253 & 29.99/0.8354 & 28.95/0.8013 & 27.42/0.8359 \\
        & SRFBN~\cite{li2019feedback} & 3,631K & 500.8G & 76.74ms & 34.70/0.9292 & 30.51/0.8461 & 29.24/0.8084 & 28.73/0.8641 \\          
        & IMDN~\cite{hui2019lightweight} & 703K & 71.7G & 5.36ms & 34.36/0.9270 & 30.32/0.8417 & 29.09/0.8046 & 28.17/0.8519 \\
        \hline
        \multirow{11}{*}{4}
        & FSRCNN~\cite{dong2016accelerating} & 13K & 4.6G & 0.67ms & 30.72/0.8660 & 27.61/0.7550 & 26.98/0.7150 & 24.62/0.7280 \\
        & FSRCNN* & 13K & 4.6G & 0.67ms & \underline{30.89}/\underline{0.8748} & \underline{27.72}/\underline{0.7599} & \underline{27.05}/\underline{0.7176} & \underline{24.76}/\underline{0.7358} \\
        & FSRCNN~(Ours) & 13K & 4.6G & 0.67ms & {\bf{30.95}}/{\bf{0.8759}} & {\bf{27.77}}/{\bf{0.7615}} & {\bf{27.08}}/{\bf{0.7188}} & {\bf{24.82}}/{\bf{0.7393}} \\
        \cline{2-9}
        & Bicubic Int.& - & - & - & 28.42/0.8104 & 26.00/0.7027 & 25.96/0.6675 & 23.14/0.6577 \\
        & DRCN~\cite{kim2016deeply} & 1,774K & 17,974.3G & 243.62ms & 31.53/0.8854 & 28.02/0.7670 & 27.23/0.7233 & 25.14/0.7510 \\
        & DRRN~\cite{tai2017image} & 297K & 6,796.9G & 57.09ms & 31.68/0.8888 & 28.21/0.7721 & 27.38/0.7284 & 25.44/0.7638 \\
        & MemNet~\cite{tai2017memnet} & 677K & 2,662.4G & 8.55ms & 31.74/0.8893 & 28.26/0.7723 & 27.40/0.7281 & 25.50/0.7630 \\
        & CARN~\cite{ahn2018fast} & 1,592K & 90.9G & 3.16ms & 32.13/0.8937 & 28.60/0.7806 & 27.58/0.7349 & 26.07/0.7837 \\
        & IDN~\cite{hui2018fast} & 591K & 34.1G & 3.08ms & 31.82/0.8903 & 28.25/0.7730 & 27.41/0.7297 & 25.41/0.7632 \\
        & SRFBN~\cite{li2019feedback} & 3,631K & 281.7G & 48.39ms & 32.47/0.8983 & 28.81/0.7868 & 27.72/0.7409 & 26.60/0.8015 \\          
        & IMDN~\cite{hui2019lightweight} & 715K & 41.1G & 4.38ms & 32.21/0.8948 & 28.58/0.7811 & 27.56/0.7353 & 26.04/0.7838 \\
        \hline
      \end{tabular}
    \end{adjustbox}
  \label{tab:general_PSNR}
\end{table}

\subsection{Results}\label{sec:result}
\subsubsection{Quantitative comparison.}
We compare in Table \ref{tab:general_PSNR} the performance of our student model with the state of the art, particularly for efficient SISR methods~\cite{dong2015image,dong2016accelerating,kim2016deeply,tai2017image,tai2017memnet,hui2018fast,ahn2018fast,li2019feedback,hui2019lightweight}. For a quantitative comparison, we report the average PSNR and SSIM~\cite{wang2004image} for upsampling factors of 2, 3, and 4, on standard benchmarks~\cite{bevilacqua2012low,zeyde2010single,martin2001database,huang2015single}. We also report the number of model parameters and operations~(MultiAdds), required to reconstruct a HR image of size~$1280\times720$, and present the average runtime of each method measured on the Set5~\cite{bevilacqua2012low} using the same machine with a NVIDIA Titan RTX GPU. From this table, we can observe two things: (1) Our student model trained with the proposed framework outperforms FSRCNN~\cite{dong2016accelerating} by a large margin, consistently for all scale factors, even both have the same network architecture. It demonstrates the effectiveness of our approach to exploiting ground-truth HR images as privileged information; (2) The model trained with our framework offers a good compromise in terms of PSNR/SSIM and the number of parameters/operations/runtimes. For example, DRCN~\cite{kim2016deeply} requires 1,774K parameters, 17,974.3G operations and average runtime of 233.93ms to achieve the average PSNR of 30.75dB on Urban100~\cite{huang2015single} for a factor of 2. On the contrary, our framework further boosts FSRCNN without modifying the network architecture, achieving the average PSNR of 30.24dB with 13K parameters/6.0G operations only, while taking 0.83ms for inference.

\begin{figure}[t]
  \begin{minipage}{0.5\linewidth}
        \captionsetup{font={small}}
        \captionof{table}{Quantitative results of student networks using other SR methods. We report the average PSNR for different scale factors (2$\times$, 3$\times$, and 4$\times$) on Set5~\cite{bevilacqua2012low} and B100~\cite{martin2001database}. *: models reproduced by ourselves using the DIV2K~\cite{timofte2017ntire} dataset; Ours: student networks of our framework.}
        \begin{adjustbox}{width=1\columnwidth,center} 
          \centering
          \begin{tabular}{c|c|c|c}
          \hline
          Methods        & \begin{tabular}[c]{@{}c@{}}2x\\ Set5/B100\end{tabular} & \begin{tabular}[c]{@{}c@{}}3x\\ Set5/B100\end{tabular} & \begin{tabular}[c]{@{}c@{}}4x\\ Set5/B100\end{tabular} \\ \hline
          FSRCNN-L*      & 37.59/31.90 & 33.76/28.81 & 31.47/27.29 \\
          FSRCNN-L (Ours) & \bf{37.65}/\bf{31.92} & \bf{33.85}/\bf{28.83} & \bf{31.52}/\bf{27.30} \\ \hline
          VDSR~\cite{kim2016accurate}    & 37.53/31.90 & 33.67/28.82 & 31.35/27.29 \\
          VDSR*          & 37.64/31.96 & 33.80/28.83 & 31.37/27.25 \\
          VDSR (Ours)    & \bf{37.77}/\bf{32.00} & \bf{33.85}/\bf{28.86} & \bf{31.51}/\bf{27.29} \\ \hline
          IDN~\cite{hui2018fast}   & 37.83/32.08 & 34.11/28.95 & 31.82/27.41 \\
          IDN*           & 37.88/32.12 & 34.22/29.02 & {\bf{32.03}}/27.49 \\
          IDN (Ours)     & \bf{37.93}/\bf{32.14} & \bf{34.31}/\bf{29.03} & 32.01/\bf{27.51} \\ \hline
          CARN~\cite{ahn2018fast}  & 37.76/32.09 & 34.29/29.06 & 32.13/27.58 \\
          CARN*          & 37.75/32.02 & 34.08/28.94 & 31.77/27.44 \\ 
          CARN (Ours)    & {\bf{37.82}}/\bf{32.08} & {\bf{34.10}}/\bf{28.95} & {\bf{31.83}}/\bf{27.45} \\ \hline
          \end{tabular}
        \end{adjustbox}
        \label{tab:other-models}
  \end{minipage}\hfill
  \begin{minipage}{0.46\linewidth}
      \captionsetup{font={small}}
  	  \centering
        \includegraphics[width=1.0\textwidth]{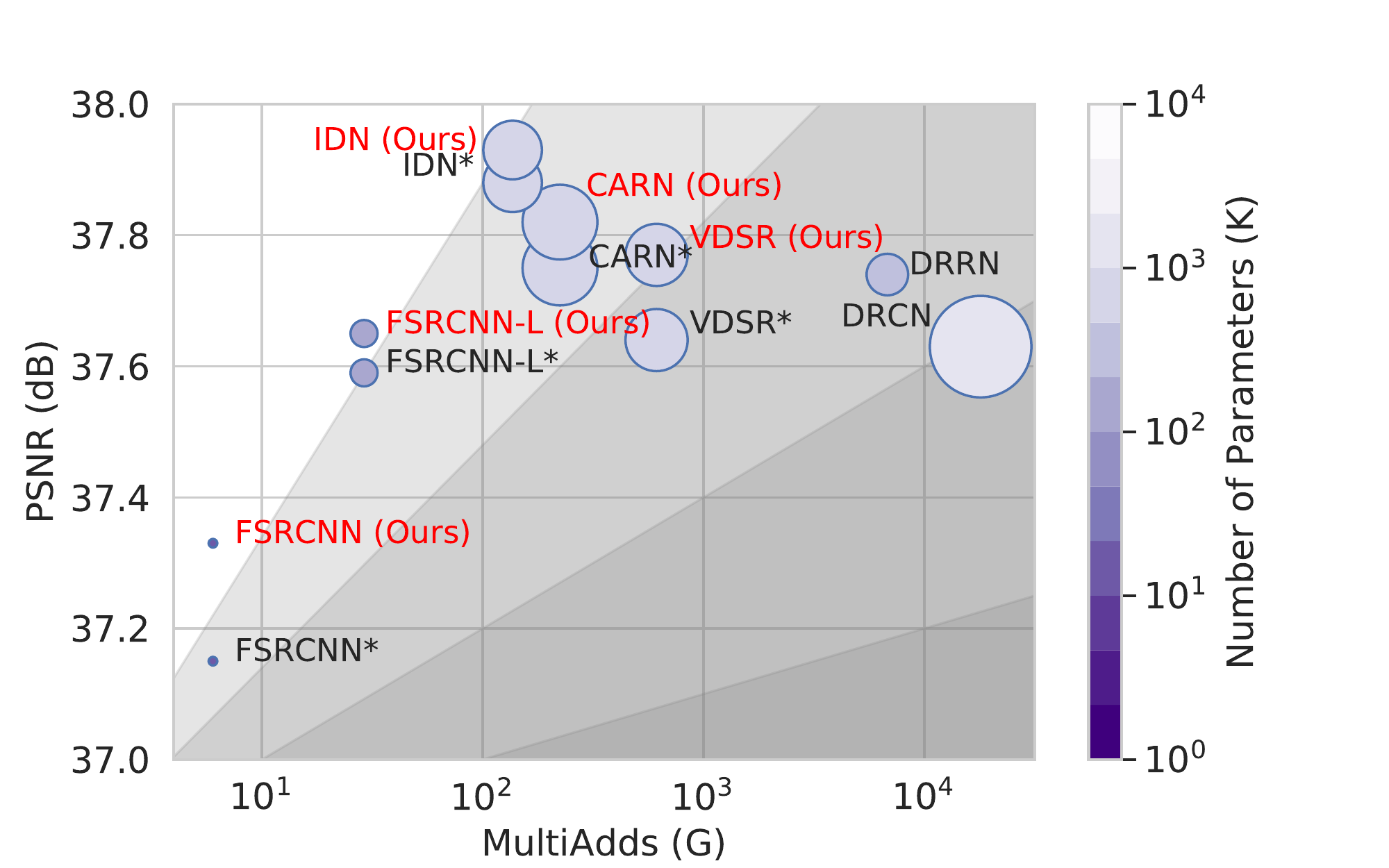}
      \caption{Trade-off between the number of operations and the average PSNR on Set5~\cite{bevilacqua2012low}~(2$\times$). The size of the circle and background color indicate the number of parameters and the efficiency of the model~(white:~high, black:~low), respectively. (Best viewed in color.)}
      \label{fig:test}
  \end{minipage}
\end{figure}

\begin{figure}[t]
  \captionsetup[subfigure]{aboveskip=1pt,belowskip=1pt,justification=centering}

  \begin{adjustbox}{width=0.93\columnwidth,center} % FSRCNN 3x urban100 71th
    \subcaptionbox*{Urban100\\ img-71 (3x)}
      {\includegraphics[width=0.250\textwidth, height=0.250\textwidth]{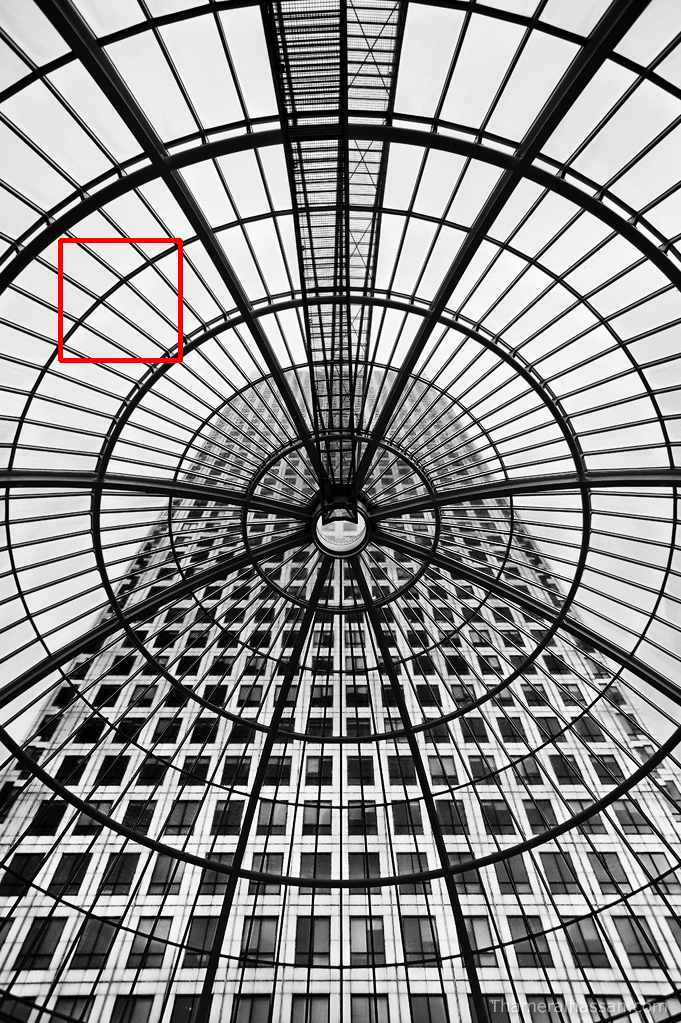}}
    \subcaptionbox*{Ground truth\\(PSNR/SSIM)}
      {\includegraphics[width=0.250\textwidth]{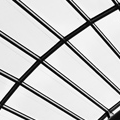}}
    \subcaptionbox*{Bicubic Int.\\(18.06/0.6835)}
      {\includegraphics[width=0.250\textwidth]{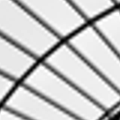}}
    \subcaptionbox*{FSRCNN$^*$\\(20.42/0.8278)}
      {\includegraphics[width=0.250\textwidth]{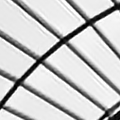}}
    \subcaptionbox*{FSRCNN~(Ours)\\(20.68/0.8416)}
      {\includegraphics[width=0.250\textwidth]{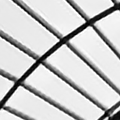}}
  \end{adjustbox}
  
  \begin{adjustbox}{width=0.93\columnwidth,center} % VDSR 2x Urban100 11th
    \subcaptionbox*{Urban100\\img-11 (2x)}
      {\includegraphics[width=0.250\textwidth, height=0.250\textwidth]{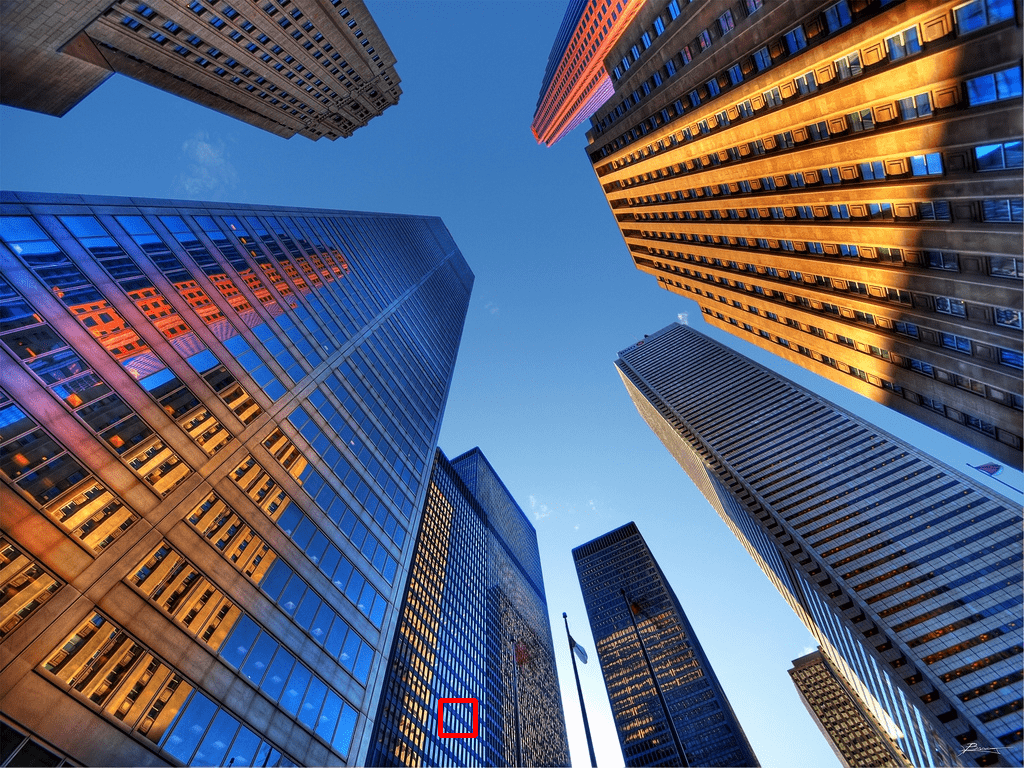}}
    \subcaptionbox*{Ground truth\\(PSNR/SSIM)}
      {\includegraphics[width=0.250\textwidth]{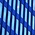}}
    \subcaptionbox*{Bicubic Int.\\(25.32/0.8034)}
      {\includegraphics[width=0.250\textwidth]{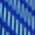}}
    \subcaptionbox*{VDSR$^*$\\(27.64/0.8993)}
      {\includegraphics[width=0.250\textwidth]{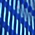}}
    \subcaptionbox*{VDSR~(Ours)\\(27.87/0.9025)}
      {\includegraphics[width=0.250\textwidth]{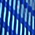}}
  \end{adjustbox}

  \begin{adjustbox}{width=0.93\columnwidth,center} % IDN 3x Set14 1st
    \subcaptionbox*{Set14\\img-1 (3x)}
      {\includegraphics[width=0.250\textwidth, height=0.250\textwidth]{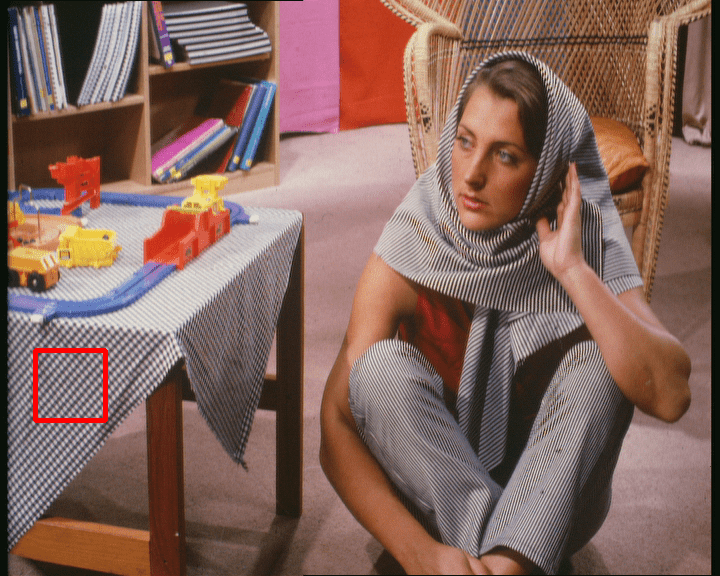}}
    \subcaptionbox*{Ground truth\\(PSNR/SSIM)}
      {\includegraphics[width=0.250\textwidth]{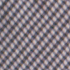}}
    \subcaptionbox*{Bicubic Int.\\(26.25/0.7538)}
      {\includegraphics[width=0.250\textwidth]{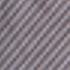}}
    \subcaptionbox*{IDN$^*$\\(26.28/0.7887)}
      {\includegraphics[width=0.250\textwidth]{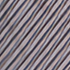}}
    \subcaptionbox*{IDN~(Ours)\\(26.97/0.8017)}
      {\includegraphics[width=0.250\textwidth]{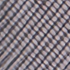}}
  \end{adjustbox}

  \begin{adjustbox}{width=0.93\columnwidth,center} % CARN 3x urban100 91th
    \subcaptionbox*{Urban100\\img-91 (3x)}
      {\includegraphics[width=0.250\textwidth, height=0.250\textwidth]{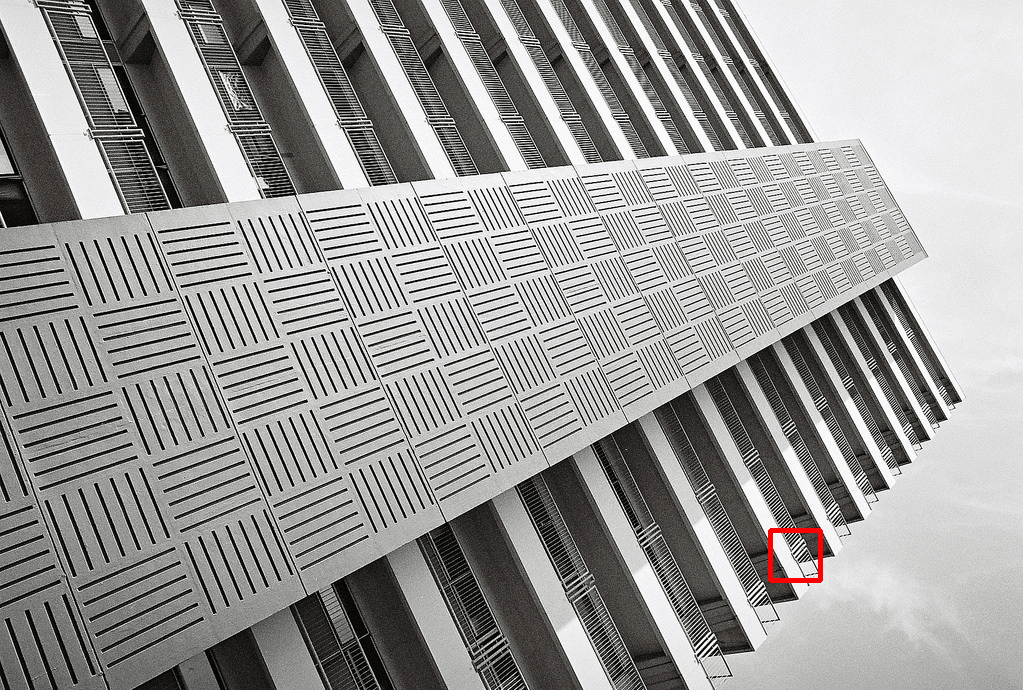}}
    \subcaptionbox*{Ground truth\\(PSNR/SSIM)}
      {\includegraphics[width=0.250\textwidth]{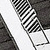}}
    \subcaptionbox*{Bicubic Int.\\(17.32/0.5164)}
      {\includegraphics[width=0.250\textwidth]{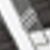}}
    \subcaptionbox*{CARN$^*$\\(20.20/0.7236)}
      {\includegraphics[width=0.250\textwidth]{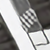}}
    \subcaptionbox*{CARN~(Ours)\\(20.32/0.7293)}
      {\includegraphics[width=0.250\textwidth]{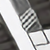}}
  \end{adjustbox}

  \vfill
  \captionsetup{font={small}}
  \caption{Visual comparison of reconstructed HR images~(2$\times$ and 3$\times$) on Urban100~\cite{huang2015single} and Set14~\cite{zeyde2010single}. We report the average PSNR/SSIM in the parentheses. (Best viewed in color.)}
  \label{fig:qualitative}    
\end{figure}

In Table~\ref{tab:other-models}, we show the performances of student networks, adopting the architectures of other SR methods, trained with our framework using the DIV2K dataset~\cite{timofte2017ntire}. We reproduce their models (denoted by *) using the same training setting but without distillation. The FSRCNN-L has the same components as FSRCNN~\cite{dong2016accelerating} but with much more parameters~(126K vs. 13K), where the numbers of filters in feature extraction and shrinking components are both 56, and the mapping module consists of 4 blocks of convolutional layers. Note that the multi-scale learning strategy in the CARN~\cite{ahn2018fast} is not used for training the network, and thus the performance is slightly lower than the original one. We can see that all the SISR methods benefit from our framework except for IDN~\cite{hui2018fast} for the scale factor of 4 on Set5. In particular, the performances of the variant of FSRCNN~\cite{dong2015image} and VDSR~\cite{kim2016accurate} are significantly boosted through our framework. Additionally, our framework further improves the performances of the cost-effective SR methods~\cite{hui2018fast,ahn2018fast}, which are specially-designed to reduce the number of parameters and operations while improving the reconstruction performance. Considering the performance gains of recent SR methods, the results are significant, demonstrating the effectiveness and generalization ability of our framework. For example IDN~\cite{hui2018fast} and SRFBN~\cite{li2019feedback} outperform the second-best methods by 0.05dB and 0.02dB, respectively, in terms of PSNR on Set5~\cite{bevilacqua2012low} for a factor of 2. We visualize in Fig.~\ref{fig:test} the performance comparison of student networks using various SR methods and the state of the art in terms of the number of operations and parameters. It confirms once more the efficiency of our framework. 

\subsubsection{Qualitative results.}
We show in Fig.~\ref{fig:qualitative} reconstruction examples on the Urban100~\cite{huang2015single} and Set14~\cite{zeyde2010single} datasets using the student networks. We can clearly see that the student models provide better qualitative results than their baselines. In particular, our models remove artifacts~(\eg,~the borders around the sculpture in the first row) and reconstruct small-scale structures~(\eg,~windows in the second row and the iron railings in the last row) and textures~(\eg,~the patterns of the tablecloth in the third row). More qualitative results can be seen in the supplementary material.

\section{Conclusion}
We have presented a novel distillation framework for SISR leveraging ground-truth HR images as privileged information. The detailed analysis on each component of our framework clearly demonstrates the effectiveness of our approach. We have shown that the proposed framework substantially improves the performance of FSRCNN as well as other methods. In future work, we will explore distillation losses specific to our model to further boost the performance.

\subsection*{Acknowledgement.}
This research was supported by the Samsung Research Funding \& Incubation Center for Future Technology (SRFC-IT1802-06).

\bibliographystyle{splncs04}
\bibliography{main}



\section*{Supplementary material}

# Learning with Privileged Information for Efficient Image Super-Resolution Supplement

Wonkyung Lee*, Junghyup Lee*, Dohyung Kim*, and Bumsub Ham†

Yonsei University

In this supplementary material, we present an analysis of different sizes of the scale maps and the imitation loss in Sec. 1 and 2, respectively, and an ablation study on our framework including comparisons on a large dataset in Sec. 3. We also show more qualitative results of the student networks which have the same architectures as various SR methods in Sec. 4.

Table A: Average PSNR of student networks (FSRCNN [4]) with different sizes of the scale maps on the Set5 [3] dataset (2×).

| Dimension of scale parameter | PSNR |
|---|---|
| $\mathbb{R}^{1 \times 1 \times 1}$ | 37.29 (+0.14) |
| $\mathbb{R}^{1 \times H' \times W'}$ | 37.29 (+0.14) |
| $\mathbb{R}^{C \times 1 \times 1}$ | 37.32 (+0.17) |
| $\mathbb{R}^{C \times H' \times W'}$ | **37.33** (+0.18) |

Table B: Average PSNR of student networks (FSRCNN [4]) trained with different teacher networks using various balance parameters $\lambda^{\mathcal{T}}$ on the Set5 [3] dataset (2×).

| $\lambda^{\mathcal{T}}$ | PSNR |
|---|---|
| 0 | 37.23 (+0.05) |
| $10^{-6}$ | 37.25 (+0.10) |
| $10^{-4}$ | **37.33** (+0.18) |
| $10^{-2}$ | 37.27 (+0.12) |
| 1 | 37.25 (+0.10) |

## 1 The size of scale map

We show in Table A performance comparisons of student networks having the same architecture as FSRCNN [4] for different sizes of scale maps in terms of the average PSNR. It shows that we can achieve the best performance, when the sizes of the scale map and a feature map in the teacher are the same. This adaptively controls the extent of distillation on each element of the feature map in the teacher network.

## 2 Imitation loss

We show in Table B performance comparisons of student networks having the same architecture as FSRCNN [4]. They are trained with different teacher networks that use different parameters $\lambda^{\mathcal{T}}$ for the imitation loss. The imitation loss encourages compact features $\hat{X}^{\mathcal{T}}$ extracted from HR images to be close to the LR counterparts, facilitating the initialization of a student network. We can see from the first row that the performance gain decreases without the imitation loss, suggesting that this loss is crucial for the weight transfer in our framework.

* equal contribution    † corresponding author (bumsub.ham@yonsei.ac.kr)

Table C: Average PSNR of student networks, trained with variants of our framework, on the B100 [11] dataset.

| Hourglass architecture | Weight transfer | $L_{im}^{\mathcal{T}}$ | $L_{distill}^{\mathcal{S}}$ | Student PSNR |
|:---:|:---:|:---:|:---:|:---:|
| - | - | - | - | 31.58 (baseline) |
| ✗ | - | - | MAE | 31.60 (+0.02) |
| ✓ | ✗ | ✗ | MAE | 31.61 (+0.04) |
| ✓ | ✓ | ✗ | MAE | 31.63 (+0.05) |
| ✓ | ✓ | ✓ | MAE | 31.65 (+0.07) |
| ✓ | ✓ | ✓ | $VID_G$ [2] | **31.66** (+0.08) |
| ✓ | ✓ | ✓ | $VID_L$ [2] | 31.65 (+0.07) |

Table D: Average PSNR of student networks with different distillation frameworks and losses on the Set5 [3] and B100 [11] datasets.

| Type | Teacher | | $L_{distill}^{\mathcal{S}}$ | Student PSNR | |
|:---:|:---:|:---:|:---:|:---:|:---:|
| | Model | Input | | Set5 [3] | B100 [11] |
| Knowledge Distillation | FSRCNN-L | LR | MAE | 37.20 (+0.05) | **31.61** (+0.03) |
| | FSRCNN-L | LR | FitNet [12] | 37.16 (+0.01) | 31.59 (+0.01) |
| | FSRCNN-L | LR | AT [14] | 37.21 (+0.06) | **31.61** (+0.03) |
| | FSRCNN-L | LR | SRKD [6] | 37.21 (+0.06) | 31.60 (+0.02) |
| | FSRCNN-L | LR | $VID_L$ [2] | **37.23** (+0.08) | **31.61** (+0.03) |
| Proposed Distillation | Ours | HR | MAE | 37.27 (+0.12) | **31.65** (+0.07) |
| | Ours | HR | FitNet [12] | 37.31 (+0.16) | 31.64 (+0.06) |
| | Ours | HR | AT [14] | 37.31 (+0.16) | **31.65** (+0.07) |
| | Ours | HR | SRKD [6] | 37.29 (+0.14) | 31.64 (+0.06) |
| | Ours | HR | $VID_L$ [2] | **37.33** (+0.18) | **31.65** (+0.07) |

If the parameter $\lambda^{\mathcal{T}}$ becomes too large (*e.g.*, 1 in the fifth row), the imitation loss forces the compact features to be identical to the LR images, which is relatively easy to achieve. In this case, our framework can be viewed as a self-distillation method [5], which however does not benefit from privilege information and degrades the performance (*e.g.*, a 0.08dB decrease in the fifth row compared to the result in the third row). Overall, the parameter $\lambda^{\mathcal{T}}$ of $10^{-4}$ gives a good compromise between imitated and privileged features, allowing the student network to achieve the best result.

## 3    Ablation studies

We present an analysis on each component of our framework using the Set5 [3] and B100 [11] datasets. We show in Table C, corresponding to Table 1 in the paper, the average PSNR on B100 [11] for student networks trained with variants of our framework. We can see that the average PSNR gradually increases by adding each component of our framework, which coincide with the findings on Set5 [3]. Table D compares the performance of the student networks using different distillation methods and losses. The first five and the last five rows show the average PSNR of the conventional knowledge distillation framework for network

compression and our framework, respectively, with different distillation losses. We can see our framework shows better performance than the knowledge distillation framework consistently in terms of PSNR on the Set5 [3] and B100 [11] datasets. This demonstrates that distilling features from the privileged information (*i.e.*, ground-truth HR images) is important, regardless of loss functions, and the proposed framework is more effective for the SISR task, boosting the performance of the student by a large margin. Note that the losses we used are typically leveraged for classification tasks to transfer 'dark knowledge' [7]. To our knowledge, there are no distillation losses effective to regression tasks. We do believe that the performance of our framework can be further boosted by the loss function specially-designed for our framework or at least for the regression tasks.

## 4   Qualitative results

Figures A, B, C, D, E show reconstruction examples on Set14 [15], B100 [11], and Urban100 [8] datasets using student networks, adopting the architectures of FSRCNN-L, FSRCNN [4], IDN [9], CARN [1], and VDSR [10], respectively. We can clearly see that the student networks trained with our framework consistently show better visual results compared with the original ones. Especially, our student networks accurately reconstruct sharp boundaries (*e.g.*, the alphabet in the last row in Fig. A and the marina floor in the first row in Fig. C), small-scale structures (*e.g.*, windows in the second and third rows in Fig. B, and the iron railings in the third row in Fig. C), textures (*e.g.*, the patterns of the zebra in the third row in Fig. D), and straight lines (*e.g.*, the ceiling of the bus stop in the second row in Fig. C and the bridge in the third row in Fig. E). These results indicate that our framework is generally applicable to the other SR methods, such as IDN, CARN, VDSR, FSRCNN, and the variant of FSRCNN.

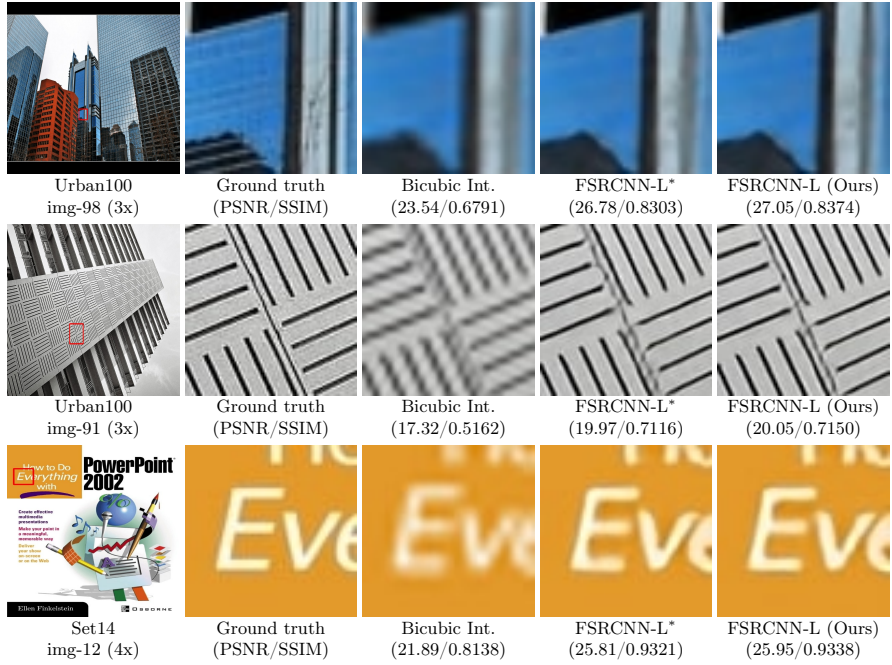

Fig. A: Visual comparison of reconstructed HR images (3× and 4×) on Set14 [15] and Urban100 [8]. We report the average PSNR/SSIM in the parentheses. Compared to the FSRCNN-L, our model reconstructs straight lines and object boundaries more accurately. *: models reproduced by ourselves using the DIV2K dataset [13] without distillation; Ours: student networks of our framework. (Best viewed in color.)

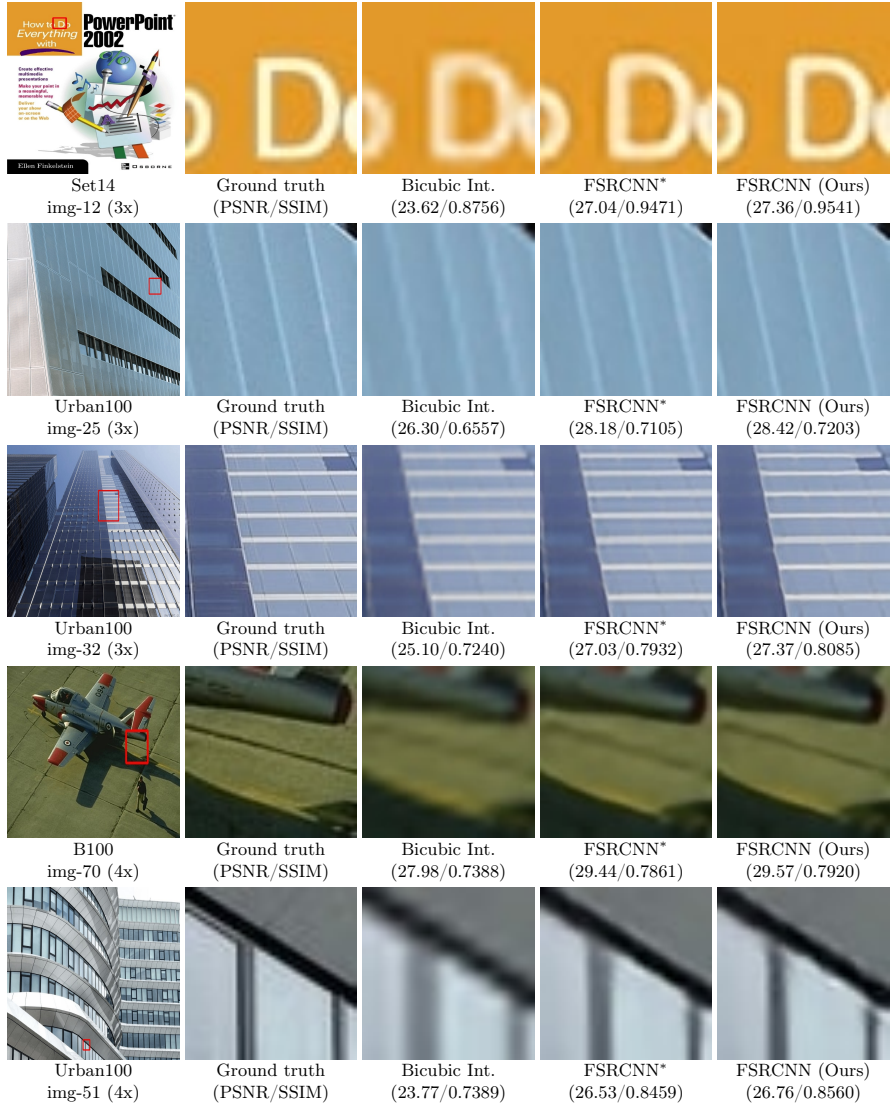

Fig. B: Visual comparison of reconstructed HR images (3× and 4×) on Set14 [15], Urban100 [8], and B100 [11]. We report the average PSNR/SSIM in the parentheses. Compared to the baseline FSRCNN, our model reconstructs small-scale structures, straight lines, and object boundaries more accurately. (Best viewed in color.)

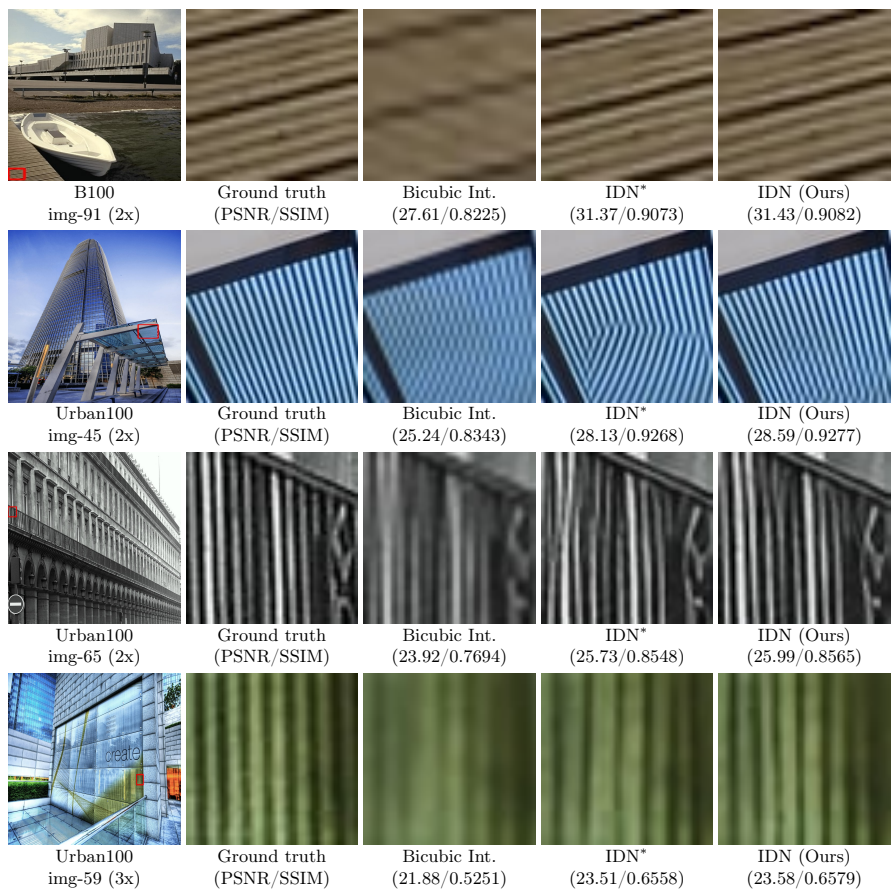

Fig. C: Visual comparison of reconstructed HR images (2× and 3×) on B100 [11] and Urban100 [8]. We report the average PSNR/SSIM in the parentheses. Compared to the baseline IDN, our model reconstructs repetitive patterns, straight lines, and object boundaries more accurately. (Best viewed in color.)

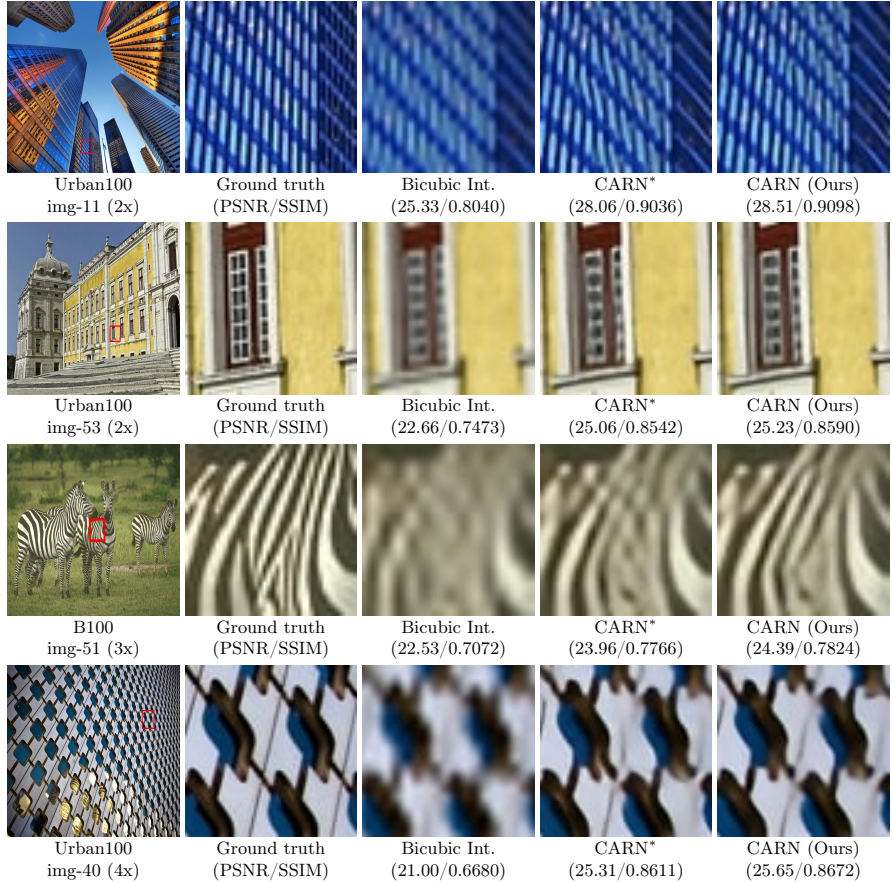

Fig. D: Visual comparison of reconstructed HR images (2×, 3×, and 4×) on Urban100 [8] and B100 [11]. We report the average PSNR/SSIM in the parentheses. Compared to the baseline CARN, our model reconstructs small-scale structures, repetitive patterns, and straight lines more accurately. (Best viewed in color.)

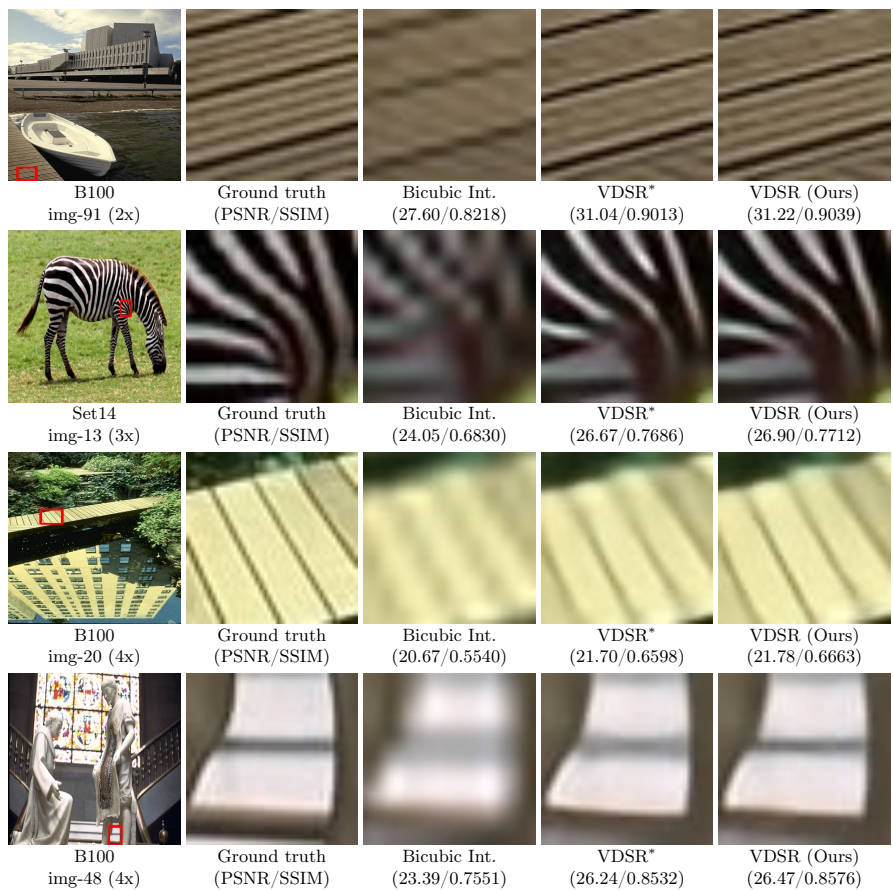

Fig. E: Visual comparison of reconstructed HR images (2×, 3×, and 4×) on B100 [11] and Ser14 [15]. We report the average PSNR/SSIM in the parentheses. Compared to the baseline VDSR, our model reconstructs small-scale structures, straight lines, and object boundaries more accurately. (Best viewed in color.)


\end{document}